%% file: paper.tex
\documentclass[sigconf]{acmart}

\usepackage{color}
\usepackage{colortbl}
\usepackage{xcolor}
\usepackage{multirow}
\usepackage{graphicx}
\usepackage{subfigure}

\newcommand{\ie}{\emph{i.e.,~}}
\newcommand{\etal}{\emph{et al.~}}
\newcommand{\eg}{\emph{e.g.,~}}

\AtBeginDocument{%
  \providecommand\BibTeX{{%
    \normalfont B\kern-0.5em{\scshape i\kern-0.25em b}\kern-0.8em\TeX}}}

\setcopyright{acmcopyright}
\copyrightyear{2021}
\acmYear{2021}

\copyrightyear{2021} 
\acmYear{2021} 
\setcopyright{acmlicensed}\acmConference[MM '21]{Proceedings of the 29th ACM International Conference on Multimedia}{October 20--24, 2021}{Virtual Event, China}
\acmBooktitle{Proceedings of the 29th ACM International Conference on Multimedia (MM '21), October 20--24, 2021, Virtual Event, China}
\acmPrice{15.00}
\acmDOI{10.1145/3474085.3475418}
\acmISBN{978-1-4503-8651-7/21/10}

\begin{document}
\fancyhead{} 
\title{Multi-Modal Multi-Instance Learning for Retinal Disease Recognition}


\author{Xirong Li}
\affiliation{
  \institution{MoE Key Lab of DEKE, \\Renmin University of China}
  \country{China}
  }
\email{xirong@ruc.edu.cn}

\author{Yang Zhou}
\affiliation{%
  \institution{Vistel AI Lab, \\ Beijing Visionary Intelligence Ltd.}
  \country{China}
}
\email{yang.zhou@vistel.cn}

\author{Jie Wang}
\author{Hailan Lin}
\affiliation{%
  \institution{AIMC Lab, School of Information, \\ Renmin University of China}
  \country{China}
}

\author{Jianchun Zhao}
\author{Dayong Ding}
\affiliation{%
  \institution{Vistel AI Lab, \\ Beijing Visionary Intelligence Ltd.}
  \country{China}
}

\author{Weihong Yu}
\author{Youxin Chen}
\affiliation{%
  \institution{Dept. of Ophthalmology, Peking Union Medical College Hospital}
  \country{China}
}

\renewcommand{\shortauthors}{Li et al.}

\begin{abstract}
This paper attacks an emerging challenge of multi-modal retinal disease recognition. Given a multi-modal case consisting of a color fundus photo (CFP) and an array of OCT B-scan images acquired during an eye examination, we aim to build a deep neural network that recognizes multiple vision-threatening diseases for the given case. As the diagnostic efficacy of CFP and OCT is disease-dependent, the network's ability of being both selective and interpretable is important. Moreover, as both data acquisition and manual labeling are extremely expensive in the medical domain, the network has to be relatively lightweight for learning from a limited set of labeled multi-modal samples. Prior art on retinal disease recognition focuses either on a single disease or on a single modality, leaving multi-modal fusion largely underexplored. We propose in this paper Multi-Modal Multi-Instance Learning (MM-MIL) for selectively fusing CFP and OCT modalities. Its lightweight architecture (as compared to current multi-head attention modules) makes it suited for learning from relatively small-sized datasets. For an effective use of MM-MIL, we propose to generate a pseudo sequence of CFPs by over sampling a given CFP. The benefits of this tactic include well balancing instances across modalities, increasing the resolution of the CFP input, and finding out regions of the CFP most relevant with respect to the final diagnosis. Extensive experiments on a real-world dataset consisting of 1,206 multi-modal cases  from 1,193 eyes of 836 subjects demonstrate the viability of the proposed model.
\end{abstract}




\begin{teaserfigure}
\centering
\subfigure[\textbf{A multi-modal sample acquired during an eye examination}, consisting of a color fundus photo (CFP) and an array of 12 OCT B-scan images.]{
\centering
\includegraphics[width=0.9\columnwidth]{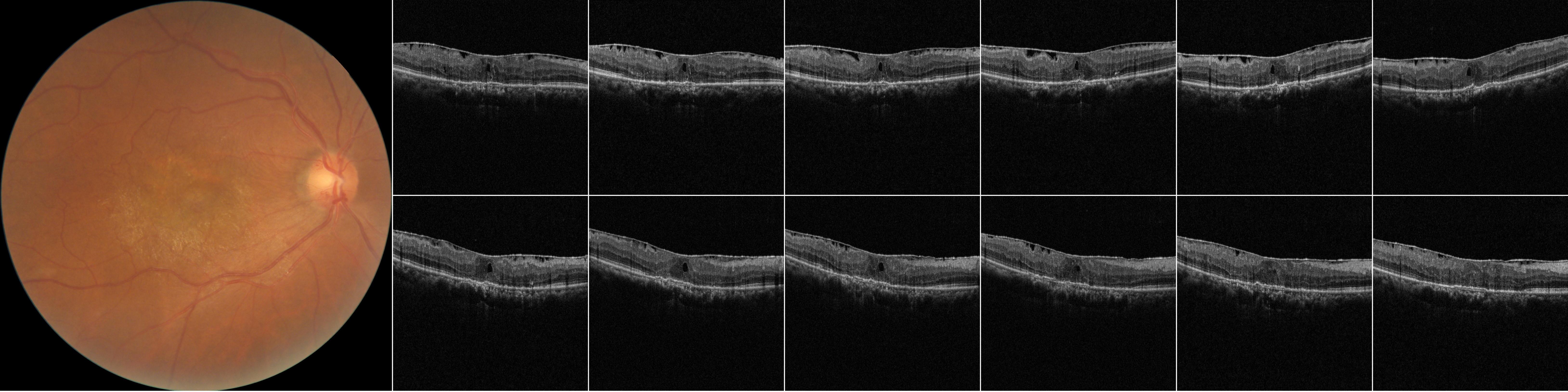}
}
\subfigure[\textbf{Activation maps produced by our proposed model}, showing abnormal regions in the CFP and abnormal B-scans in the OCT array.]{
\centering
\includegraphics[width=0.9\columnwidth]{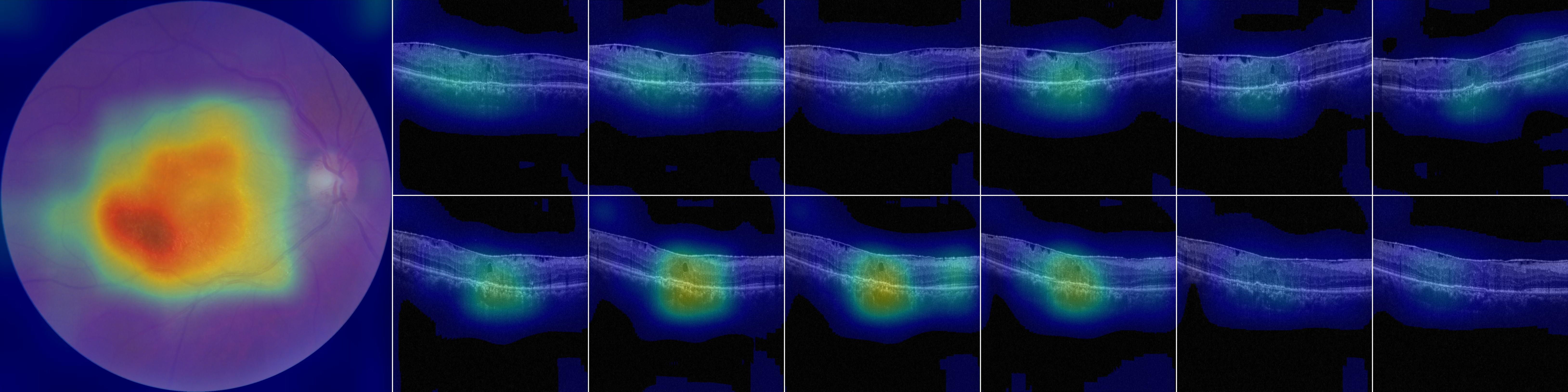}
}
\vspace{-3mm}
\caption{\textbf{Illustrating multi-modal retinal disease recognition}, with (a) input and (b) output. Diseases: ERM, ME.} 
\label{fig:showcase} 
\end{teaserfigure}

\maketitle

\section{Introduction} \label{sec:intro}

\input{intro}

\section{Related Work} \label{sec:relatd}

\input{related}
\section{Proposed Method} \label{sec:method}

\input{method}

\section{Evaluation} \label{sec:eval}

\input{eval}

\section{Conclusions} \label{sec:conc}

We develop an end-to-end deep learning approach to multi-modal retinal disease recognition. Extensive experiments on a real-world dataset support the following conclusions. As the efficacy of color fundus photography and OCT scans is disease-dependent, the ability of being both selective and interpretable is important for multi-modal fusion. The proposed MM-MIL module possesses both properties, as demonstrated by its superior performance against the prior art. Moreover, MM-MIL has substantially fewer parameters than the prevalent Multi-Head self-Attention module, and thus can be trained on relatively small-sized data. All this makes MM-MIL attractive for AI-assisted retinal disease diagnosis.

\medskip
\textbf{Acknowledgements}. This research was supported by BJNSF (4202033), BJNSFC Haidian Original Innovation Joint Fund (19L2062), the Pharmaceutical Collaborative Innovation Research Project of Beijing Science and Technology Commission (Z191100007719002), and Public Computing Cloud, Renmin University of China.
\bibliographystyle{ACM-Reference-Format}
\balance
\bibliography{sample-base}


\end{document}

%% file: intro.tex
Multi-modal imaging is a routine measure in the screening and diagnosis of many diseases. Consider retinal disease diagnosis for instance.  Retinal diseases manifest themselves in various ways in retina -- a ten-layered structure covering over 200 degrees of eye fundus. Accurate diagnoses depend on accurate localization and visualization of the pathological alterations in this subtle and complex structure.
Images of distinct modalities are acquired during an eye examination, including a color fundus photograph (CFP) and an array of optical coherence tomography (OCT) B-scan images, see Fig. \ref{fig:showcase}. 
OCT B-scans provide cross-sectional information of the retina and choroid, while CFPs show en face information of the fundus. Due to such different imaging mechanisms, using a specific modality alone is often inadequate to spot alterations for all diseases. For dryAMD at its early stage, drusenoid pigment epithelial detachment can be observed in OCT images, even when its counterpart remains invisible in a CFP. Similarly for Macular Edema (ME), its main symptom is retinal thickening. The resultant alteration in CFP is hard exudate (not obvious in the early stage) and that in OCT B-scan is cystoid macular edema (noticeable as the layered structure shown in B-scan changes tangibly). Hence, OCT B-scans play a crucial role in recognizing ME, dryAMD and wetAMD in particular at their early stage. Meanwhile, as typical alterations by DR include microaneurysm, retinal hemorrhage, and venous beading, all are vascular lesions, and thus more observable in CFP than in OCT B-scans, making CFP more suitable for DR recognition. For EpiRetinal Membrane (ERM) and Pathological Myopia (PM), both modalities matter. In clinical practice, ophthalmologists use both modalities as a standard method for fundus checks. 
Hence, multi-modal deep learning that effectively exploits a limited set of labeled multi-modal samples is crucial for artificial intelligence (AI) assisted disease diagnosis. 

Previous efforts on retinal disease recognition are mostly based on a single modality, let it be a CFP \cite{bjo2021-mdd}, an OCT image \cite{mmm20-wu-oct} or a sequence of OCT images \cite{deepmind-oct}. They are thus not directly applicable to deal with the multi-modal input. An initial attempt on end-to-end multi-modal learning is by Wang \etal \cite{miccai19-amd}, where authors present MM-CNN, a two-stream CNN that takes a CFP and a single OCT image as a paired input. While MM-CNN is shown to be superior to its single-modal counterparts for categorizing subclasses of age-related macular degeneration (AMD), the OCT image has to be manually selected by a technician that operates the OCT device. We argue that this requirement limits the practical use of MM-CNN, as the technician is unlikely to be sufficiently trained to select the most appropriate B-scan images for all diseases. Moreover, by using only the chosen image with the other B-scans ignored, the OCT modality is underexplored. 

Multi-modal CNNs that accept a sequence of images as input have been extensively studied in the context of video action classification \cite{cvpr17-twostream,hara3dcnns,icmr15-eval-two-stream}. To exploit the spatio-temporal information in video frames, a 3D-CNN consisting of stacked 3D convolutional blocks is commonly used, either alone \cite{hara3dcnns} or in a two-stream manner where another 3D-CNN runs in parallel to process the sequence of optical flow images \cite{cvpr17-twostream,icmr15-eval-two-stream}. Note that the good performance of 3D-CNNs is subject to the availability of very large-scale training data \cite{cvpr17-twostream}. This is however difficult to be fulfilled in the medical domain where both data acquisition and manual labeling are expensive. Moreover, as the sequence is processed as a whole, 3D-CNNs lack an explicit mechanism to interpret the contribution per frame, a wanted property for AI-assisted disease diagnosis. Therefore, multi-modal CNNs with a relatively lightweight architecture and interpretability is in demand for retinal disease recognition.

For automated screening of Retinopathy of Prematurity (ROP) given multiple CFPs acquired per case, Li \etal \cite{icpr20-rop} adapt instance-attention based deep multiple instance learning (MIL) \cite{icml18-mil}. The given CFPs, treated as a bag of instances, are fed in parallel into a shared 2D-CNN for feature extraction. Such instance-level features are weighed in terms of the corresponding attended weights and summed up for case-level classification. The shared 2D backbone substantially reduces the amount of trainable parameters, whilst the attention mechanism naturally explains the contribution of each CFP to the final prediction. Again, as \cite{icpr20-rop} works on a single modality, how to exploit MIL for the multi-modal scenario remains open. In fact, because the CFP is clearly outnumbered by the OCT images in a given input, extending MIL by treating the CFP and OCT images as individual instances in a bag is problematic.

Inspired by the success of deep MIL for the ROP classification task, we propose in this paper Multi-Modal Multi-Instance Learning (MM-MIL) for multi-label retinal disease recognition. The overall architecture, as illustrated in Fig. \ref{fig:model}, follows the classical two-stream framework \cite{cvpr17-twostream}, with MM-MIL as a novel module for multi-modal feature fusion. To conquer the data imbalance issue and to improve the resolution of the CFP channel, over sampling on CFP is performed to form a pseudo sequence of CFPs. In sum, our contributions are as follows:
\begin{itemize}
\item We propose MM-MIL, a new module for multi-modal feature fusion. It inherits the interpretability of the instance-attention based MIL. Its lightweight architecture (as compared to current multi-head attention modules \cite{nips2017-transformer}) makes it suited for learning from relatively small-sized datasets.
\item For an effective use of MM-MIL, we propose to generate a pseudo sequence of CFPs by over sampling a given CFP. The benefits of this tactic are multifold, including well balancing the samples between the CFP and OCT modalities, increasing the resolution of the CFP input, and finding out regions of the CFP most relevant with respect to the final prediction. 
\item Extensive experiments on a real-world dataset collected from an outpatient clinic justify the superiority of the proposed method against the state-of-the-art. Compared to the best single-modal baseline (OCT-MIL with an overall AP of 0.7748) and the best multi-modal baseline (MM-CNN++ with AP of 0.8172), our model scores the highest AP of 0.8539. 
\end{itemize}

%% file: related.tex
Since the seminal work by Google \cite{jama16-dr}, which shows the initial success of deep CNNs for CFP-based diabetic retinopathy (DR) screening, many deep learning based methods have been proposed for eye disease recognition \cite{jamaoph17-amd,ophalmology17-dr,icpr20-rop,icme20-detect-pm,mmm2021-glaucoma,mm20-cataract,cell2018-oct}. The majority of the methods target at a single disease, \eg DR \cite{ophalmology17-dr}, Pathological Myopia  (PM) \cite{icme20-detect-pm}, Glaucoma \cite{mmm2021-glaucoma} or AMD \cite{or17-amd}, and make predictions based on a single-modal input,  either a CFP or an OCT image. 

Although scarce, there are emerging efforts on recognizing multiple diseases from a CFP \cite{bjo2021-mdd} or from a sequence of OCT B-scan images \cite{deepmind-oct}. Li \etal \cite{bjo2021-mdd} train an ensemble of SeResNext-50 to detect twelve major retinal diseases. To process OCT volume data, De Fauw \etal \cite{deepmind-oct} first train a 3D-Unet to segment layers of B-scan images, and then feed the segmentation results into a 3D-DenseNet for multiple disease diagnosis and referral recommendation. Again, the above methods are single-modal.



Probably due to the lack of public multi-modal data, we see few work on multi-modal retinal disease recognition. In a pilot study \cite{miccai19-amd}, Wang \etal develop a two-stream CNN termed MM-CNN for multi-modal AMD categorization. Given a CFP and a manually chosen OCT B-scan image as a paired input, MM-CNN uses a ResNet-18 to extract features from the CFP and another ResNet-18 to extract features from the OCT image. The CFP and OCT feature vectors are simply concatenated and fed into a regular classification block (a linear layer followed by softmax) for categorizing the input as normal, dryAMD, or wetAMD. Despite its encouraging performance against the single-modal alternatives, we argue that MM-CNN has the following deficiencies. First, manual selection of one B-scan introduces extra workload on the technician. It also affects the model performance when there is discrepancy between the B-scan selection strategy used in the test stage and that used for training, see our experiments. Moreover, using merely one image per OCT scan makes the OCT modality largely underexplored. Lastly, fusion by feature concatenation lacks the ability of selectively exploiting the multi-modal information for different diseases. 





Technically, we are inspired by the successful use of instance-attention based deep multiple instance learning (MIL) \cite{icml18-mil} for ROP classification \cite{icpr20-rop}. In order to capture different zones of the premature retina, an ROP examination typically collects multiple CFPs. By viewing each CFP as an instance, Li \etal \cite{icpr20-rop} formulate ROP classification as an MIL problem. As their study is single-modal, how to exploit MIL for the multi-modal scenario is untouched.  Straightforward solutions such as combining MM-CNN with MIL, \eg substituting MIL for the OCT branch in MM-CNN, are inadequate for multi-modal retinal disease recognition.

The state-of-the-art on multi-modal representation, \eg cross-attention \cite{cvpr20-wei} and co-attention \cite{mm20-cheng}, is essentially multi-head self-attention (MHA) used in Transformer-alike architectures \cite{nips2017-transformer}. Compared to MHA, our proposed MM-MIL performs better for the new task yet with much less parameters. Moreover, while the interpretability of Transformers is a research problem on its own \cite{cvpr21-transformer-interp}, MM-MIL has an explicit mechanism for model interpretability.

%% file: method.tex
\begin{figure*}[tbh]
\centering
\includegraphics[width=\textwidth]{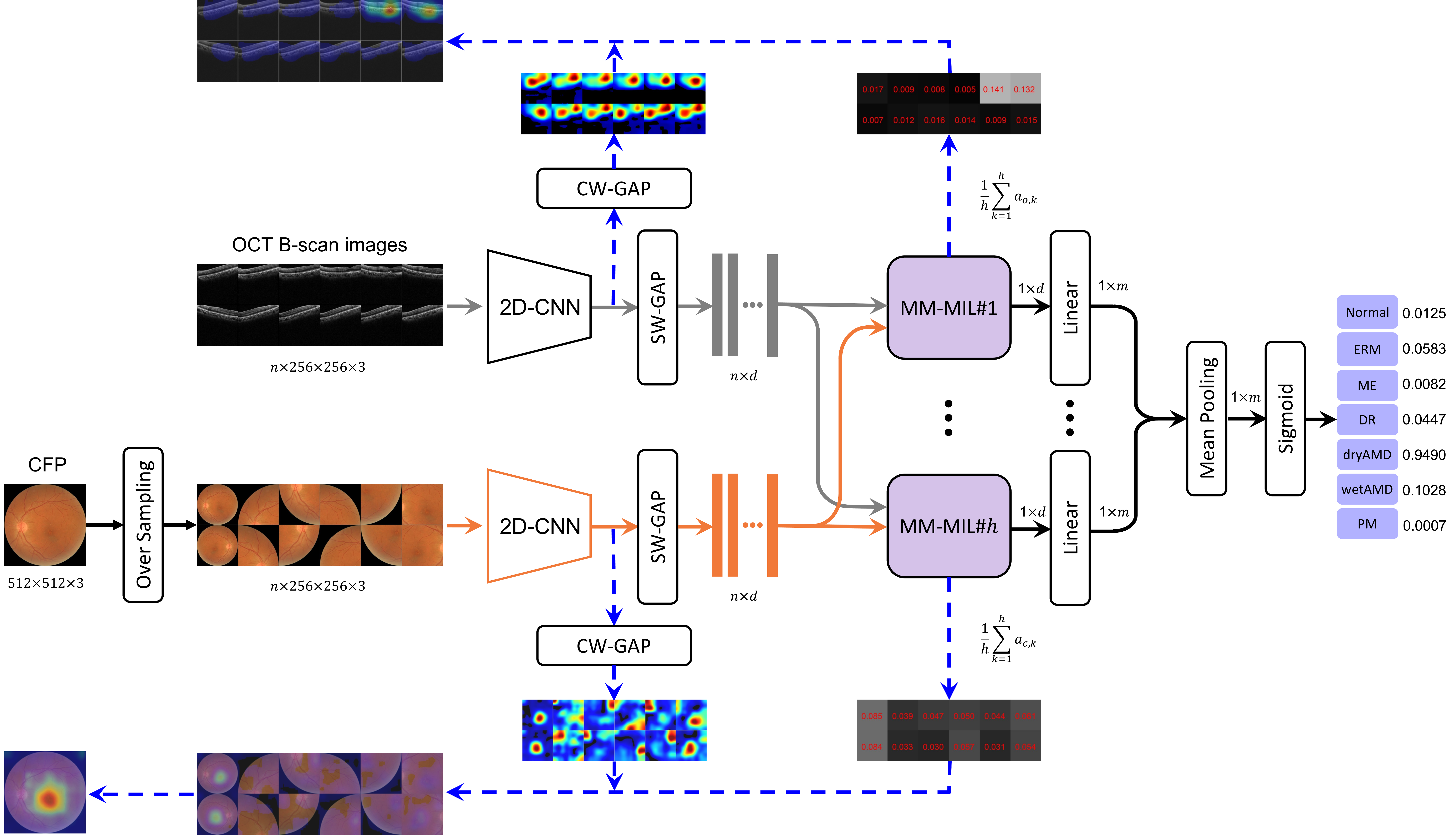}
\caption{\textbf{Proposed multi-modal retinal disease classification network in its inference mode}. Given a multi-modal case with a CFP and a sequence of $n$ OCT B-scan images, the network performs over sampling on the given CFP to generate a sequence of $n$ CFPs. The two sequences are then fed in parallel into two 2D-CNNs to extract feature maps of size $w\times h \times d$ per instance. Per modality, spatial-wise global average pooling (SW-GAP) is applied upon the feature maps to obtain a $d$-dimensional feature vector for each instance. The proposed Multi-Modal Multiple Instance Learning (MM-MIL) module with $h$ heads, see Fig. \ref{fig:mm-mil}, aggregates the $2n$ instance-level features into $h$ case-level features, each of which is converted into category-wise decision scores by a linear layer. Final probabilistic prediction is obtained by mean pooling plus sigmoid activation. For model interpretation, the feature maps per instance are converted to a $w\times h$ activation map by channel-wise global average pooling (CW-GAP). Multiplied by the corresponding instance-attention weight from MM-MIL, the activation maps are used to visualize the contribution of the OCT images and regions of the CFP to the prediction. Best viewed on screen.} 
\label{fig:model}
\end{figure*}

Based on multi-modal fundus imaging comprised of OCT and color fundus photography, we aim for automated categorization of the fundus condition of a specific eye. We use $x$ to indicate an eye. Its examination by a multi-modal fundus camera produces a color fundus photograph (CFP), denoted as $x_c$, and an array of $n$ OCT B-scan images, denoted as $\mathbf{x}_o = \{x_{o,1}, \ldots, x_{o,n}\}$. Suppose there are $m$ distinct categories to be considered. We aim to build a deep neural network $G$ that accepts the multi-modal input $(x_c, \mathbf{x}_o)$ and produces a probabilistic vector $p$ accordingly, 
\begin{equation}\label{eq:basic}
p := G(x_c, \mathbf{x}_o),
\end{equation}
where $p_i$ denotes the probability of the eye belonging to class $i$, $i=1,\ldots,m$. 
In what follows, we describe the overall architecture of $G$ in Section \ref{ssec:overall-network}, followed by MM-MIL in Section \ref{ssec:mm-mil} and model interpretation in Section \ref{ssec:model-interpret}. 

\subsection{Multi-Modal Deep Classification Network} \label{ssec:overall-network}

The overall architecture is illustrated in Fig. \ref{fig:model}. Our proposed network conceptually consists of three blocks, \ie 1) instance-level feature extraction, 2) multi-modal feature fusion, and 3) case-level classification.

\textbf{1) Instance-level feature extraction}. In order to match the number of the instances in the OCT modality, we perform over sampling on the original CFP to produce a sequence of $n$ (sub-)images, denoted as $\{x_{c,1}, \ldots, x_{c,n}\}$. Specifically, the multi-modal fundus camera used in our experiments ran in a radial scan mode, producing $n=12$ B-scans per case. Therefore, in the test stage, we crop the four corners and the center of the original CFP with a fixed-sized window. With horizontal flip on the big image and the five sub-images, the over sampling operation eventually generates 12 images in total for the CFP modality. 

Due to the noticeable difference in the visual appearance of CFP and OCT, we use two 2D-CNNs to extract instance-level features for the two modalities, respectively.  The backbone of both 2D-CNNs is ResNet-50 \cite{cvpr2016-resnet}, initialized using ImageNet-pretrained models. Per OCT instance $x_{o,i}$, we use $F_{o,i}$ to denote the feature maps of size $w \times h \times d$ extracted by the OCT 2D-CNN. In a similar vein, we define $F_{c,i}$ for a CFP instance $x_{c,i}$. By applying spatial-wise global average pooling (SW-GAP) on the feature maps, we obtain $n$ $d$-dimensional instance-level features per modality, denoted as $\{f_{o,1}, \ldots, f_{o,n}\}$ and $\{f_{c,1}, \ldots, f_{c,n}\}$, respectively.  

\textbf{2) Multi-modal feature fusion}. As shown in Fig. \ref{fig:model}, the $2n$ instance-level features are simultaneously fed into our proposed MM-MIL module, which aggregates these features into a case-level feature $\bar{f}_{mm}$ of the same dimension $d$, expressed as 
\begin{equation}
    \bar{f}_{mm} := \mbox{MM-MIL}(\{f_{c,1}, \ldots, f_{c,n}, f_{o,1}, \ldots, f_{o,n}\}).
\end{equation}

\textbf{3) Case-level classification}. We use a linear layer to convert $\bar{f}_{mm}$ into categorize-wise decision scores. Since a specific eye might have multiple diseases, we formulate the retinal disease recognition task as a multi-label classification problem. As such, the sigmoid activation is adopted. Accordingly, the generic formula given in Eq. \ref{eq:basic} can now be realized as 
\begin{equation} \label{eq:case-clf}
    p := \mbox{sigmoid}(Linear_{d\times m}(\bar{f}_{mm})).
\end{equation}
When a multi-head MM-MIL is used to produce multiple $\bar{f}_{mm}$, we let each $\bar{f}_{mm}$ go through a distinct linear layer and use mean pooling in advance to the sigmoid layer, see Fig. \ref{fig:model}. A standard BCE loss between $p$ and case-level labels is computed  for model training. 

Next, we detail the procedure of computing the case-level modal-fused feature $\bar{f}_{mm}$.

\subsection{MM-MIL for Multi-Modal Feature Fusion} \label{ssec:mm-mil}

To make the paper self-contained, we outline the single-modality MIL block previously used for CFP-based ROP classification \cite{icpr20-rop}. Given $\{f_i\}$ as $n$ feature vectors extract from $n$ CFP instances,  instance-attention weights $\{a_i\}$ is obtained by stacking the features and feeding them into the following feedforward network,
\begin{equation} \label{eq:single-modal-mil}
\{a_i\} := \mbox{softmax}(Linear_{128 \times 1}(\mbox{tanh}(Linear_{d\times 128}(\{f_i\})))).
\end{equation}
Accordingly, a case-level feature $\bar{f}$ is obtained as a weighted sum of the instance-level features, \ie 
\begin{equation} \label{eq:weighted-sum}
    \bar{f} := \sum_{i=1}^n a_i f_i.
\end{equation}

\begin{figure}[tbh]
    \centering
    \includegraphics[width=0.9\columnwidth]{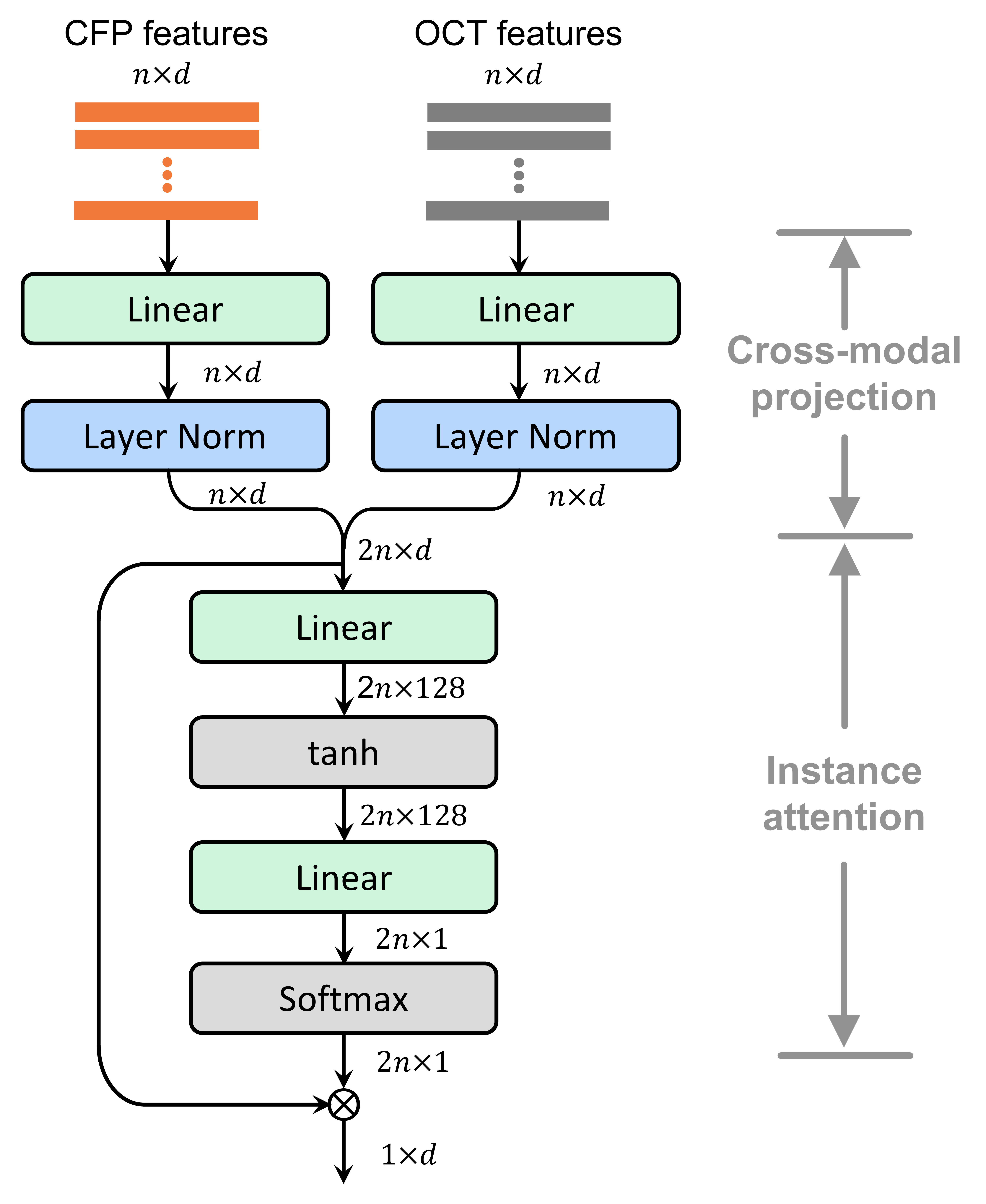}
    \caption{\textbf{Proposed Multi-Modal Multiple Instance Learning (MM-MIL) module for multi-modal feature fusion}. Instance-level CFP / OCT features are aggregated with instance-attention weights into a case-level feature vector. A multi-head version of MM-MIL can be obtained with ease by using multiple instance-attention blocks. }
    \label{fig:mm-mil}
\end{figure}

In the multi-modal context, as the CFP features $\{f_{c,i}\}$ and OCT features $\{f_{o,i}\}$ are extracted by distinct 2D-CNNs, they are not comparable by definition. In order to make them additive, we introduce a cross-modal projection block, implemented as a modality-specific linear layer followed by Layer Normalization \cite{layer-norm}, see Fig. \ref{fig:mm-mil}. Consequently, the CFP features are transformed into new features of the same size $\{\hat{f}_{c,i}\}$, \ie 
\begin{equation} \label{eq:projection}
    \hat{f}_{c,i} := \mbox{LayerNorm}(Linear_{d\times d}(f_{c,i})).
\end{equation}
In a similar manner we obtain the new OCT features $\{\hat{f}_{o,i}\}$.

By stacking $\{\hat{f}_{c,i}\}$ and $\{\hat{f}_{o,i}\}$ and substituting them for $\{f_i\}$ in Eq. \ref{eq:single-modal-mil}, we obtain instance-attention weights for the $2n$ multi-modal instances, denoted by $\{a_{c,1}, \ldots, a_{c,n}, a_{o,1}, \ldots, a_{o,n}\}$. By definition, we have $\sum_{i=1}^n a_{c,i} + \sum_{i=1}^n a_{o,i}=1$, where the first term indicates the importance of the CFP modality and the second term indicates the importance of the OCT modality. The case-level modal-fused  feature is computed by putting the corresponding terms into Eq. \ref{eq:weighted-sum}:
\begin{equation}
    \bar{f}_{mm} := \sum_{i=1}^n a_{c,i} f_{c,i} + \sum_{i=1}^n a_{o,i} f_{o,i}.
\end{equation}

A conceptual diagram of MM-MIL (with a single head) is shown in Fig. \ref{fig:mm-mil}. By using multiple instance-attention blocks, a multi-head MM-MIL can be derived with ease. As Table \ref{tab:mmmil_vs_mha} shows, when compared to multi-head self-attention (MHA), the proposed MM-MIL has much fewer parameters. 

\input{tables/attn_params}

\subsection{Model Interpretability} \label{ssec:model-interpret}

As aforementioned, the contribution of the individual instances is subject to their attention weights $\{a_{c,1}, \ldots, a_{c,n}, a_{o,1},\ldots,a_{o,n}\}$ calculated by MM-MIL. Meanwhile, the feature maps of each instance reflect how it responses spatially. Therefore, combining the two factors enables the interpretability of our multi-modal classification network in multiple aspects. In particular, for the OCT modality, sorting the B-scans by their attention weights in descending order helps select the most abnormal B-scan for further examination. For the CFP modality, as each instance corresponds to a specific region of the original CFP, overlaying an attention-weight based heatmap on top of the CFP visualizes which region is the most relevant w.r.t. the final prediction. Moreover, comparing the accumulated attention weights per modality reveals which modality the model truly counts on for recognizing specific diseases. 

For a specific instance, say $x_{c,i}$, we compute its attention-weighed activation map $A_{c,i}$ as
\begin{equation} \label{eq:activation-map}
    A_{c,i} := a_{c,i} \cdot \mbox{CW-GAP}(F_{c,i}),
\end{equation}
where CW-GAP indicates channel-wise global average pooling that compresses $F_{c,i}$ into a $w\times h$ feature map. For an MM-MIL with $h$ heads, $a_{c,i}$ is computed by $\frac{1}{h}\sum_{k=1}^h a_{c,k,i}$, where $a_{c,k,i}$ is the attention weight of $x_{c,i}$ produced by MM-MIL\#$k$.
By default, MM-MIL refers to a four-head MM-MIL, \ie MM-MIL$\times 4$, unless otherwise stated. 

%% file: tables/attn_params.tex
\begin{table}[tbh]
\caption{The number of trainable parameters in distinct attention modules. The number of depth and heads in Multi-Head self-Attention (MHA) is denoted by $d$ and $h$, respectively. MM-MIL$\times h$ stands for MM-MIL with $h$ heads. Even compared with a mini-version of MHA with $d=1$ and $h=1$, the MM-MIL series has substantially fewer parameters.}
\label{tab:mmmil_vs_mha}
\begin{tabular}{|l|r|}
\hline
\textbf{Attention module} & \multicolumn{1}{c|}{\textbf{\#Parameters}}  \\ \hline
MHA($d=1$, $h=1$) & 50.35 M \\ \hline
MHA($d=4$, $h=4$) & 201.41 M \\ \hline
MM-MIL$\times$1 & 8.67 M\\ \hline
MM-MIL$\times$2 & 8.93 M\\ \hline
MM-MIL$\times$4 & 9.45 M\\ \hline
MM-MIL$\times$8 & 10.50 M\\ \hline
\end{tabular}
\end{table}

%% file: eval.tex
\subsection{Experimental Setup}

\begin{figure}[h]
    \centering
    \includegraphics[width=\columnwidth]{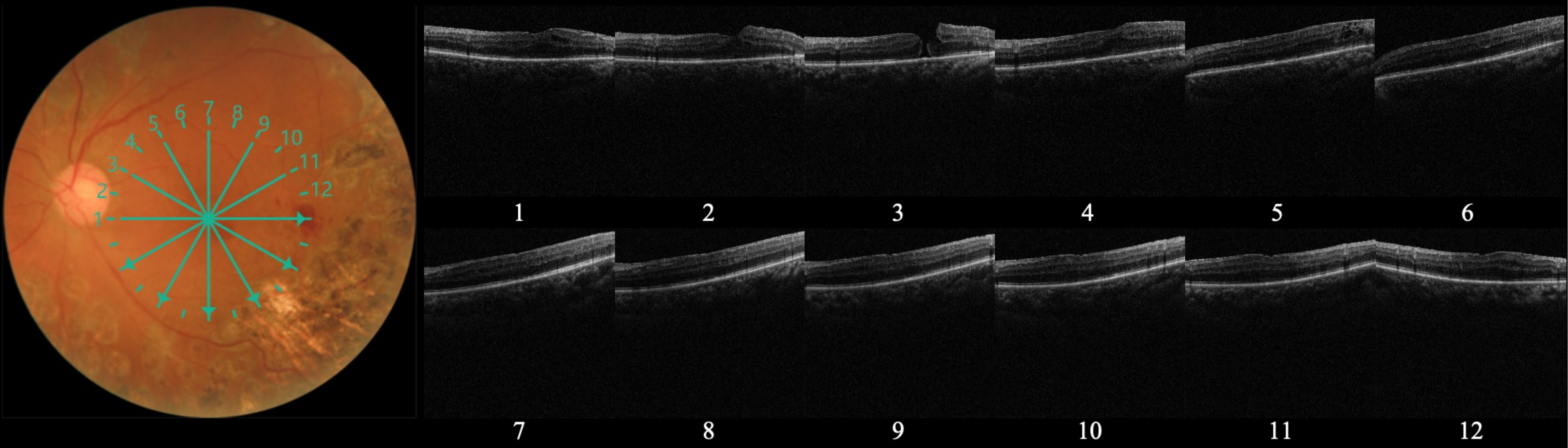}
    \caption{A multi-modal sample acquired by a Topcon Maestro-1 fundus camera in the radial scan mode. Expert labels: ERM and DR.}
    \label{fig:radial-mode}
\end{figure}

\textbf{Data collection}. Due to the lack of public data for multi-modal retinal disease recognition, we built a new dataset as follows. Multi-modal cases were collected using a  Topcon Maestro-1 (Topcon Corp., Japan) at the outpatient clinic, the Department of Ophthalmology in a state hospital from July 2020 to October 2020. The radial scan mode of the Topcon fundus camera was used, allowing us to acquire a CFP and 12 OCT B-scans simultaneously in a single examination, see Fig. \ref{fig:radial-mode}. A panel of four retinal experts was formed and asked to categorize each case with respect to dozens of pre-specified diseases. For quality control, each case was labeled by two experts independently. In case of disagreement, a third expert was asked to make the final decision. Note that not all diseases had a reasonable amount of cases for training and evaluation. Eventually, our dataset consists of 1,206 multi-modal cases from 1,193 eyes of 836 subjects, where each case has been labeled either as \textit{normal} or categories from six vision-threatening diseases including \textit{epiretinal membrane} (ERM), \textit{macular edema} (ME), \textit{diabetic retinopathy} (DR), \textit{dry age-related macular degeneration} (AMD), \textit{wet AMD} and \textit{pathological myopia} (PM). The number of diseases assigned to an abnormal case ranges from 1 to 4, with an average value of 1.44. 
%
%
Some of the diseases such as DR can co-occur in both eyes of a specific subject. Hence, to avoid any risk of data leakage, we randomly split the data collection  into three disjoint subsets, \ie training, validation and test, on the basis of subjects. A person (and his / her associated samples) is exclusively assigned to one of the three subsets. The ratio of the amount of subjects in training / validation / test is 6:2:2 approximately. Table \ref{tab:dataset} summarizes basic data statistics.

\input{tables/data}

\textbf{Performance metrics}. In a retinal disease screening program, a subject with her eye(s) predicted with higher probability of having retinal diseases shall be given higher priority in a waiting list of referrals. Hence, we report Average Precision (AP) as commonly used to assess the quality of rankings. As secondary metrics, we report AUC, Sensitivity (Sen.), Specificity (Spe.) and F1-score  of Sen. and Spe. For computing Sen. and Spe., we use the default threshold of $0.5$ to convert probabilistic output to hard labels. 


\textbf{Details of implementation}.  
We run all experiments with PyTorch~\cite{nips2019-pytorch}. For a fair comparison between all models evaluated in this study, we use the following common setup: ResNet-50 as the backbone, with its weights initialized using an ImageNet-pretrained model, SGD as the optimizer with momentum of 0.9 and weight decay of $1e-5$. The initial learning rate is 0.01, with the OneCycle \cite{isop2019-1cycle} strategy to adjust the learning rate. The maximum number of training epochs is 50, except for the CFP-only ResNet which used 100 epochs due to its relatively slow convergence speed. Per training sample, we use a smaller number of 6 OCT / CFP instances for the purposes of case-level data augmentation and saving GPU memory to use bigger batch size. In particular, random down-sampling is conducted to select the OCT instances, while random crop is used to generate CFP instances for training. Every instance is resized to $256\times 256$. In addition, traditional low-level data augmentation is performed on instances. For a more reliable evaluation, for each model we repeat the training procedure three times and select the version that maximizes AP on the validation set. 


\subsection{Comparison with State-of-the-art}

\subsubsection{Baselines} 
We compare the following state-of-the-art: \\
    $\bullet$ MM-CNN~\cite{miccai19-amd}: A two-stream CNN that accepts a CFP and a manually selected OCT B-scan image as a paired input. Multi-modal feature fusion is implemented by concatenating 2,048-dimensional ResNet-50 features of CFP and OCT.  We train MM-CNN with B-scans selected on the basis of our expert annotations. \\
    $\bullet$ OCT-MIL~\cite{icpr20-rop}: A very recent method that uses MIL to exploit a set of CFPs for ROP classification. We adopt its MIL module to perform classification based on the OCT B-scan sequence. \\
    $\bullet$ OCT-Conv3D: A 3D-CNN is used by \cite{cell2018-oct} for OCT-based retinal disease classification. However, 3D-ResNet ~\cite{hara3dcnns} failed to converge on our training data. Hence, we try an alternative solution, which first utilizes 2D-ResNet for feature extraction. Three 3D conv. blocks are then applied on the re-shaped feature maps to exploit the spatio-temporal information. Global average pooling is performed to derive a case-level feature for classification. \\
    $\bullet$ MM-CNN++: We improve MM-CNN, with its original OCT branch replaced by OCT-MIL. \\
    $\bullet$ CFP: A single-modal CNN with a CFP as its input. We include this baseline for a more  comprehensive evaluation.

\subsubsection{Results} 
The performance of MM-MIL and the baselines is given in Table \ref{tab:sota-overall}. Comparing the two modalities, they perform closely on \emph{Normal} and \emph{PM}. The two OCT models perform clearly better than CFP on \emph{ERM}, \emph{ME}, \emph{dryAMD} and \emph{wetAMD}. As for \emph{DR}, its main symptoms are caused by vascular lesions, which are more vivid on color fundus photos. This explains the clear performance gap between the OCT models and the CFP model for recognizing \emph{DR}. The result justifies the necessity of a  multi-modal approach to retinal disease recognition. 

\input{tables/sota-overall}

MM-CNN performs closely to OCT-MIL, the best single-modal model. This result shows that the better performance of multi-modal fusion cannot be taken as granted. Table \ref{tab:bscan-selection} shows that MM-CNN is sensitive to the choice of the B-scan chosen for classification. 

\input{tables/bscan-selection}

Compared to OCT-MIL, MM-CNN++ improves the overall AP from 0.7710 to 0.8172 (a relative improvement of 6.0\%), making it the best multi-modal baseline. The proposed MM-MIL outperforms this strong baseline, reaching the best AP of 0.8539. It is worth pointing out that none of the multi-modal baselines can beat the single-modal baselines for all diseases. For instance, MM-CNN++ is less effective than the CFP model on \emph{DR}. By contrast, MM-MIL consistently outperforms the best of the single-modal baselines on every category. Specifically, MM-MIL$\times 4$ detects \textit{Normal} with an AP of 0.9940 (95\% confidence interval (CI), 0.9845-1.0), and the detection performance for the other classes are as follows: ERM 0.7793 (95\% CI, 0.7048-0.8538), ME 0.9038 (95\% CI, 0.8406-0.9669), DR 0.8939 (95\% CI, 0.8281-0.9597), dryAMD 0.6220 (95\% CI, 0.5028-0.7412), wetAMD 0.9035 (95\% CI,  0.7964-1.0000), PM 0.8808 (95\% CI, 0.7638-0.9977). MM-MIL is more selective than MM-CNN++.

\subsection{Ablation Study} \label{ssec:ablation}

We now evaluate the influence of three major designs, \ie the over sampling strategy, the number of heads in MM-MIL, and alternatives to MM-MIL, on the performance of our classification network. 

\subsubsection{The influence of the over sampling strategy} 
As Table \ref{tab:os} shows, MM-MIL$\times 1$ without over sampling causes a clear performance drop, with AP decreased from 0.8097 to 0.7872. We attribute this result to the reason that without over sampling, the OCT instances become dominant during the training process. Indeed, this is also confirmed by the fact that MM-MIL$\times 1$ without over sampling performs quite closely to OCT-MIL, suggesting that the CFP modality is mostly underexplored. In addition, we try over sampling in the single-modal setting, by training CFP-MIL which uses MIL to exploit the generated sequence of CFPs. The better performance of CFP-MIL against the CFP model (0.7357 versus 0.6827 in terms of AP) further justifies the viability of the over sampling tactic.

\input{tables/over-sample}

\subsubsection{Number of heads in MM-MIL}
The performance of MM-MIL with distinct number of heads is presented in the last four rows of Table \ref{tab:sota-overall}. As we have already noted, the importance of the CFP and OCT modalities varies over diseases. MM-MIL with more heads has a larger capacity to learn better instance attention. Therefore, better performance is observed on MM-MIL$\times 4$ and MM-MIL$\times 8$, with the latter slightly worse. Recall that we have 7 categories in total, making  MM-MIL$\times 4$ sufficient for the current task. 

\begin{figure}[tbh!]
    \centering
    \subfigure[Per-category performance of each head in MM-MIL$\times 4$]{
        \includegraphics[width=0.8\columnwidth]{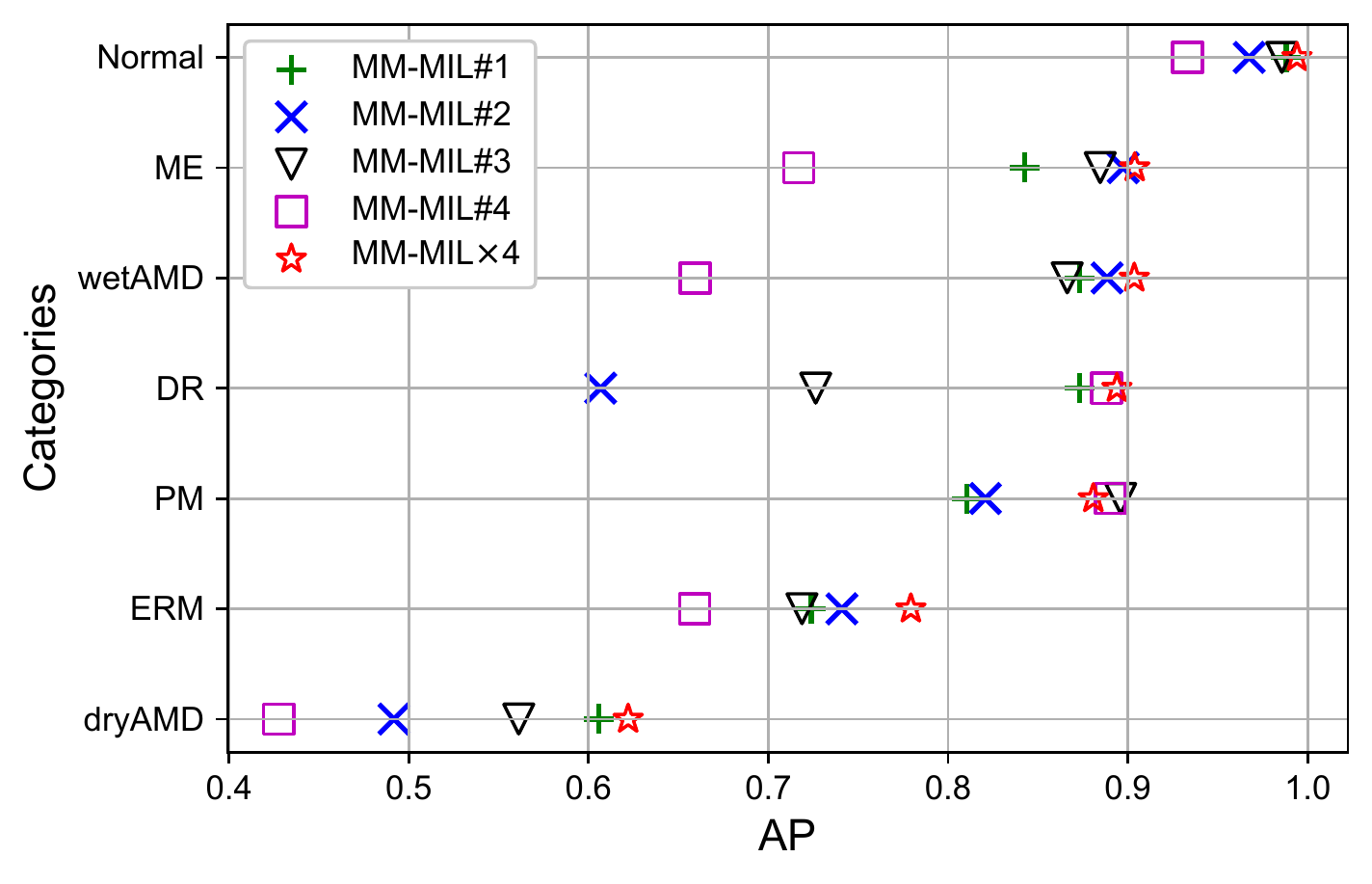}
        \label{fig:multi-head-ap}
    }
    \quad
    \subfigure[Per-category attention weights on the CFP modality]{
        \includegraphics[width=0.8\columnwidth]{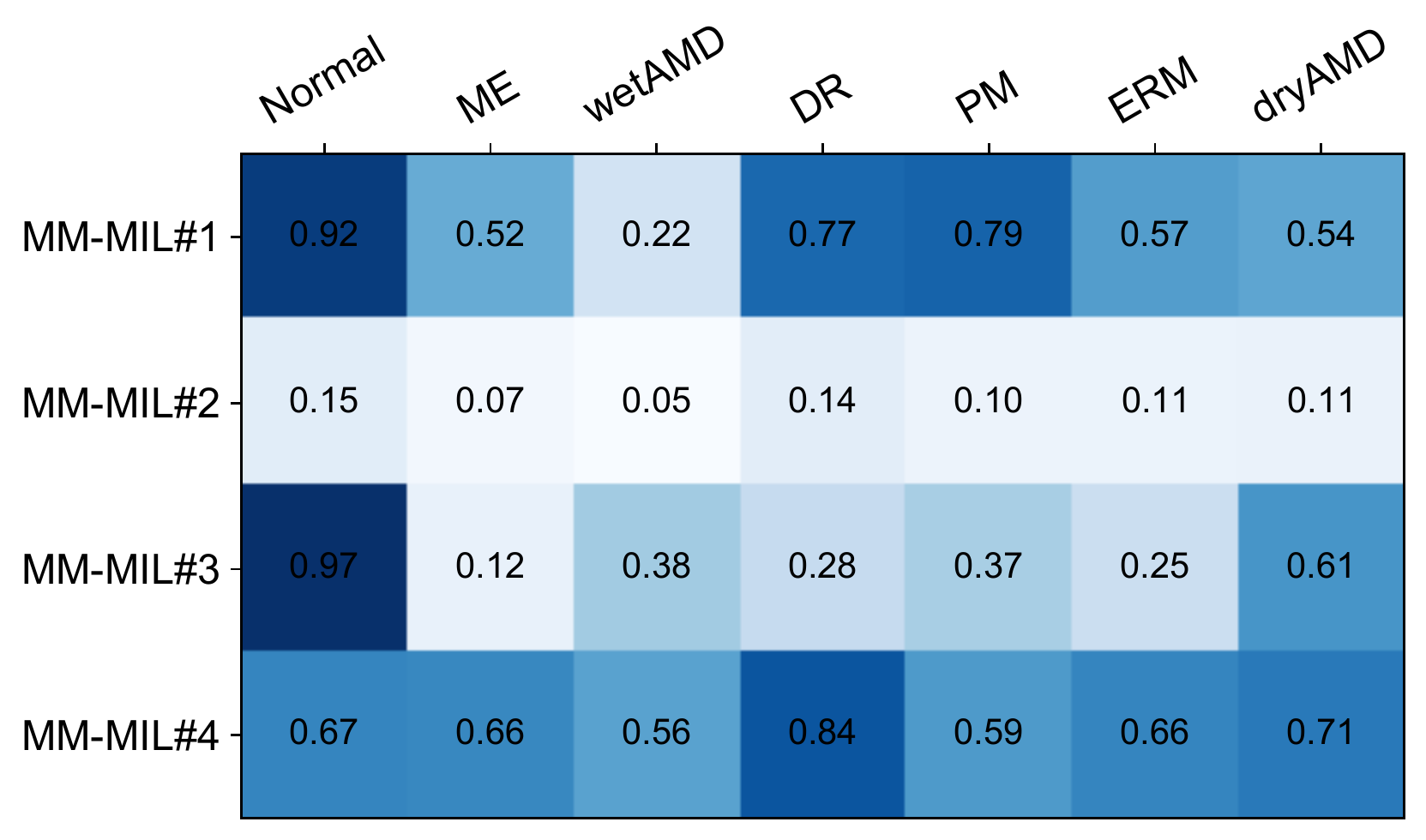}
        \label{fig:multi-head-weight}
    }
    \caption{A zoom-in view of MM-MIL$\times 4$, showing (a) per-category performance of each of the four heads and (b) how much attention the heads pay to the CFP modality. }
    \label{fig:multi-head}
\end{figure}

To better understand the behavior of MM-MIL$\times 4$, we provide a zoom-in view in Fig. \ref{fig:multi-head}. Per-category performance of each head, denoted as MM-MIL\#$k$, is plotted in Fig. \ref{fig:multi-head-ap}. Recall that the $k$-th head produces multi-modal instance attention weights. 
Per case we have $(\sum_{i=1}^n a_{c,k,i}) + (\sum_{i=1}^n a_{o,k,i})=1$. This property means for each category, the averaged value of the accumulated attention weights per modality reflects to what extent the head is attended. Accordingly, we visualize in Fig. \ref{fig:multi-head-weight} each head's per-category attention weight on the CFP modality. 

A joint use of Fig. \ref{fig:multi-head-ap} and \ref{fig:multi-head-weight} allows us to interpret the behavior of a specific head. MM-MIL\#4 puts more attention on CFP. Hence, it performs well on DR and PM for which this modality is suited, but is less effective for dryAMD and wetAMD, where lesions such as \textit{drusen} appear more vividly in OCT images. By constrast, MM-MIL\#2 pays much less attention to the CFP modality, making it the least effective for recognizing DR and PM. 



\subsubsection{MM-MIL versus MHA}



Table \ref{tab:mhsa} shows the performance of our multi-modal retinal disease classification network when substituting the popular MHA module for MM-MIL. The mini-version of MHA, which has one head and one depth only, decreases AP from 0.8539 to 0.8116. A wider and deeper MHA ($d=4$ and $h=4$) reduces the performance further. The results allow us to conclude that MHA, while being extremely popular in a wide range of tasks, is less effective than our proposed MM-MIL for multi-modal retinal disease classification.


\input{tables/mhsa}

\begin{figure}[tbh!]
    \centering
    \subfigure[Expert labels: DR $\rightarrow$ Prediction: DR (0.9542)]{
        \includegraphics[width=0.97\columnwidth]{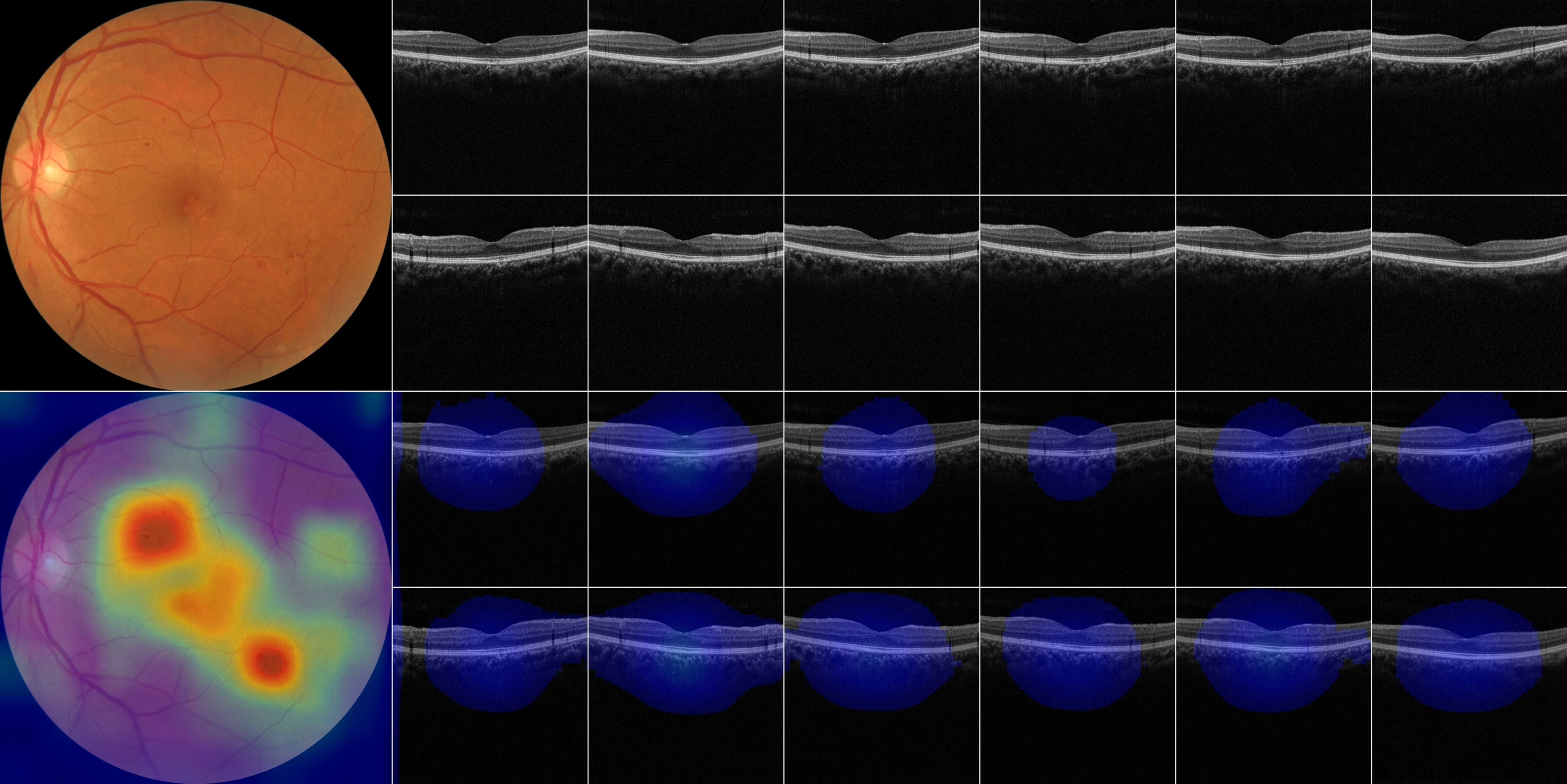}
        \label{fig:vis_dr}
    }
    \subfigure[Expert labels: wetAMD $\rightarrow$ Prediction: wetAMD (0.9967)]{
        \includegraphics[width=0.97\columnwidth]{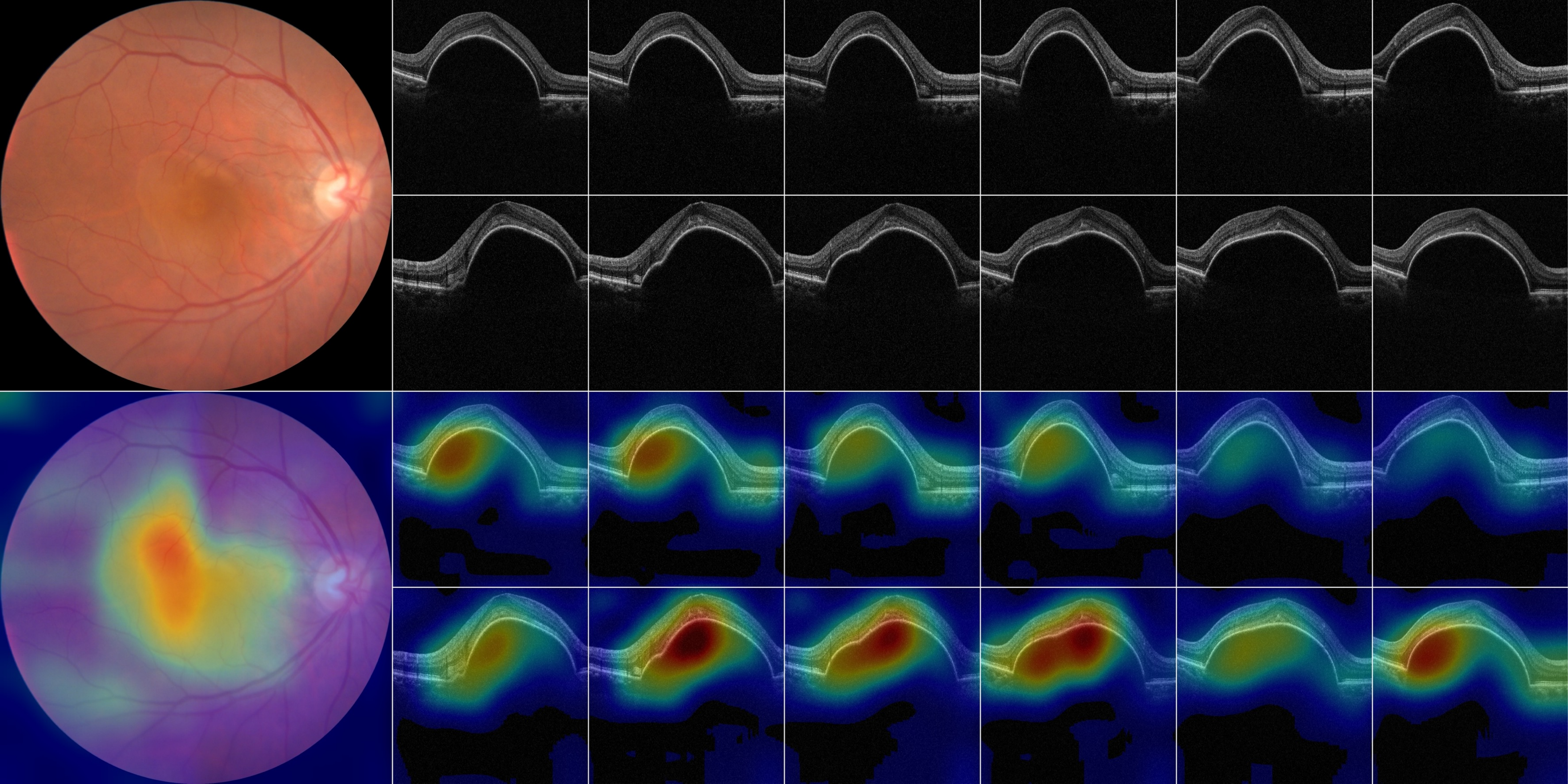}
        \label{fig:vis_wet_amd}
    }
    \subfigure[Expert labels: PM $\rightarrow$ Prediction: PM (0.9992)]{
        \includegraphics[width=0.97\columnwidth]{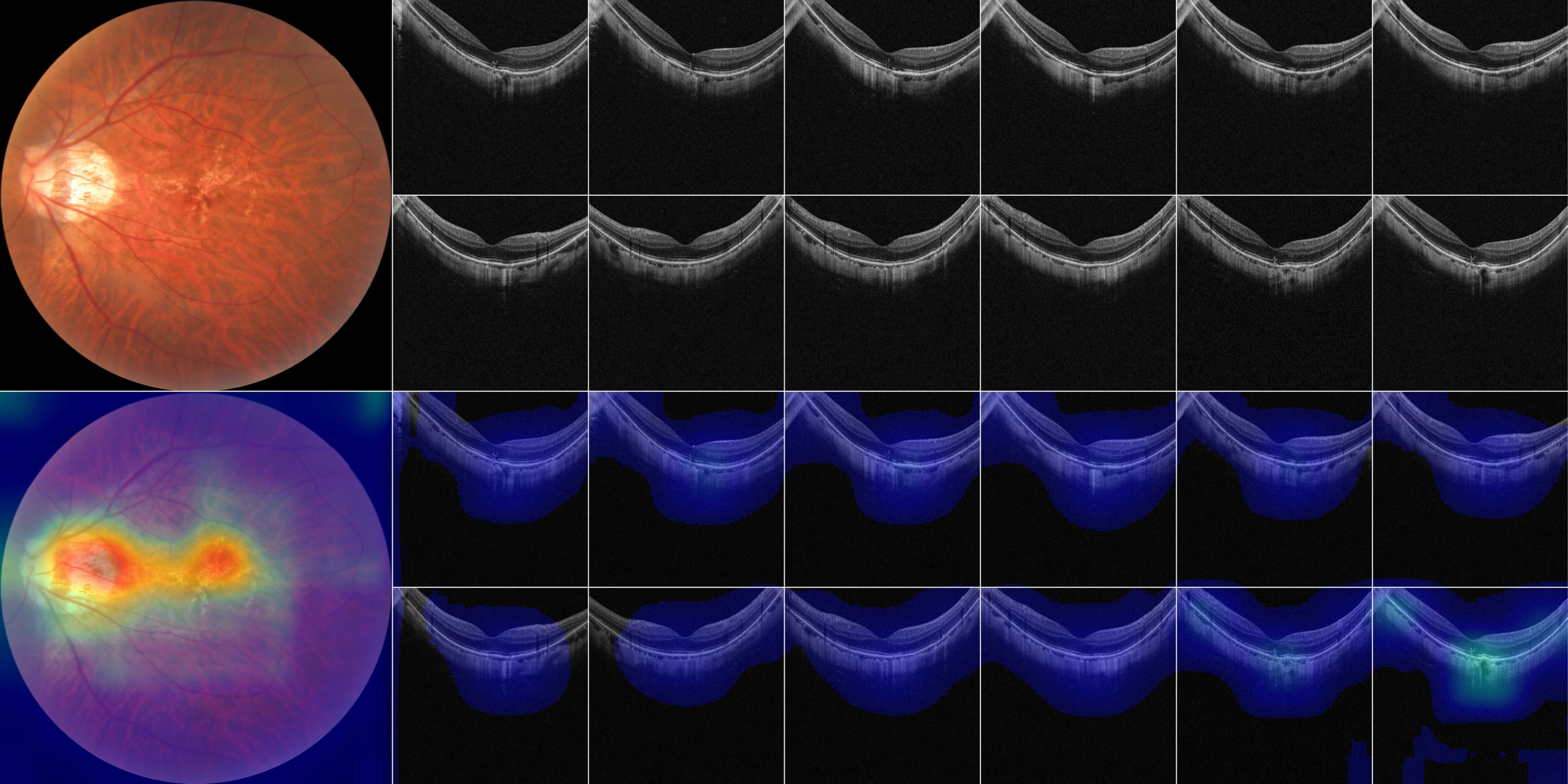}
        \label{fig:vis_pm}
    }
    \caption{Some qualitative results of model prediction and visualization. Numbers in parentheses are predicted scores. 
    }
    \label{fig:show-model-interpret}
\end{figure}

Fig. \ref{fig:show-model-interpret} shows some qualitative results of model prediction and instance attention-weighed activation maps, cf. Section \ref{ssec:model-interpret}. Consider Fig. \ref{fig:vis_pm} for instance. We observe that the activation area covers the temporal side of the optic disc (the brightest region in the CFP). This differs from the other sub-figures, neither of which has their optic disc overlapped with the activation area. Note that for PM, peripapillary atrophy is a clinical finding typically found in the area surrounding the optic disc. Recall that our model is  trained exclusively on case-level labels, the result demonstrates the viability of the activation maps  for model interpretability.

%% file: tables/data.tex
\begin{table}[htbp]
    \caption{Number of subjects, eyes and cases of each category for training, validation and test in our experiments.}
    \label{tab:dataset}
    \resizebox{\linewidth}{!}{
    \begin{tabular}{|l|r|r|r|r|r|r|r|r|r|}
    \hline
    \multirow{2}{*}{\textbf{Category}} & \multicolumn{3}{c|}{\textbf{Training}} & \multicolumn{3}{c|}{\textbf{Validation}} & \multicolumn{3}{c|}{\textbf{Test}} \\ \cline{2-10} 
     & \multicolumn{1}{l|}{\textit{Subjects}} & \multicolumn{1}{l|}{\textit{Eyes}} & \multicolumn{1}{l|}{\textit{Cases}} & \multicolumn{1}{l|}{\textit{Subjects}} & \multicolumn{1}{l|}{\textit{Eyes}} & \multicolumn{1}{l|}{\textit{Cases}} & \multicolumn{1}{l|}{\textit{Subjects}} & \multicolumn{1}{l|}{\textit{Eyes}} & \multicolumn{1}{l|}{\textit{Cases}} \\ \hline
    \textit{Normal} & 251 & 358 & 359 & 87 & 117 & 118 & 85 & 126 & 126 \\ \hline
    \textit{ERM} & 126 & 144 & 145 & 44 & 53 & 54 & 51 & 57 & 57 \\ \hline
    \textit{ME} & 95 & 104 & 105 & 30 & 35 & 35 & 36 & 41 & 41 \\ \hline
    \textit{DR} & 74 & 106 & 108 & 23 & 37 & 37 & 27 & 41 & 41 \\ \hline
    \textit{dryAMD} & 53 & 64 & 65 & 19 & 21 & 21 & 20 & 26 & 26 \\ \hline
    \textit{wetAMD} & 36 & 39 & 40 & 12 & 14 & 14 & 14 & 14 & 14 \\ \hline
    \textit{PM} & 27 & 35 & 38 & 10 & 15 & 15 & 11 & 14 & 14 \\ \hline
    Total & 485 & 700 & 710 & 173 & 238 & 241 & 178 & 255 & 255 \\ \hline
    \end{tabular}
    }
\end{table}

%% file: tables/sota-overall.tex
\begin{table*}[]
    \caption[]{\textbf{Performance of the state-of-the-art for retinal disease recognition}. Neither of the multi-modal baselines, \ie MM-CNN and MM-CNN++, can beat the best of the single-modal baselines (highlighted in lightblue cell) for all diseases. Our proposed MM-MIL$\times 4$ does so, showing its effectiveness for multi-modal feature fusion.}
    \label{tab:sota-overall}
    \resizebox{\textwidth}{!}
    {
    \begin{tabular}{|l|c|c|c|c|c|c|c|c|c|c|c|c|}
\hline
 & \multicolumn{5}{c|}{Overall performance} & \multicolumn{7}{c|}{\textbf{AP per category}} \\ \cline{2-13} 
\multirow{-2}{*}{\textbf{Model}} & \textit{AP} & \textit{AUC} & \textit{Sen.} & \textit{Spe.} & \textit{F1} & \textit{Normal} & \textit{ERM} & \textit{ME} & \textit{DR} & \textit{dryAMD} & \textit{wetAMD} & \textit{PM} \\ \hline
\multicolumn{13}{|l|}{\textit{Single-modal:}} \\ \hline
CFP & 0.6827 & 0.9025 & 0.6109 & 0.9500 & 0.7295 & \cellcolor[HTML]{6DAFD6}0.9830 & 0.5508 & 0.6405 & \cellcolor[HTML]{6DAFD6}0.8738 & 0.3879 & 0.5277 & 0.8152 \\ \hline
OCT-MIL & 0.7748 & 0.9281 & 0.6640 & 0.9495 & 0.7417 & 0.9775 & 0.7173 & \cellcolor[HTML]{6DAFD6}0.8451 & 0.7442 & 0.5045 & 0.8108 & 0.8245 \\ \hline
OCT-Conv3D & 0.7710 & 0.9283 & 0.6324 & 0.9370 & 0.7141 & 0.9756 & \cellcolor[HTML]{6DAFD6}0.7578 & 0.8390 & 0.6408 & \cellcolor[HTML]{6DAFD6}0.5049 & \cellcolor[HTML]{6DAFD6}0.8195 & \cellcolor[HTML]{6DAFD6}0.8597 \\ \hline
\multicolumn{13}{|l|}{\textit{Multi-modal:}} \\ \hline
MM-CNN & 0.7743 & 0.9072 & 0.7244 & 0.9522 & 0.8128 & 0.9761 & 0.7248 & 0.8020 & 0.7607 & 0.5210 & 0.8151 & 0.8200 \\ \hline
MM-CNN++ & 0.8172 & 0.9403 & 0.6917 & 0.9612 & 0.7892 & 0.9861 & 0.7626 & 0.8429 & 0.8409 & 0.5764 & 0.8087 & 0.9030 \\ \hline
MM-MIL$\times$ 1 & 0.8097 & 0.9261 & 0.7311 & 0.9635 & 0.8214 & 0.9888 & 0.7576 & 0.7802 & 0.8617 & 0.5326 & 0.8325 & \textbf{0.9149} \\ \hline
MM-MIL$\times$ 2 & 0.8348 & 0.9424 & \textbf{0.7577} & 0.9664 & \textbf{0.8393} & 0.9933 & 0.7628 & 0.8409 & 0.8898 & 0.5584 & 0.8969 & 0.9014 \\ \hline
MM-MIL$\times$ 4 & \textbf{0.8539} & \textbf{0.9545} & 0.7431 & 0.9643 & 0.8291 & \textbf{0.9940} & 0.7793 & \textbf{0.9038} & 0.8939 & \textbf{0.6220} & 0.9035 & 0.8808 \\ \hline
MM-MIL$\times$ 8 & 0.8534 & 0.9505 & 0.7335 & \textbf{0.9720} & 0.8285 & 0.9922 & \textbf{0.7939} & 0.8633 & \textbf{0.9165} & 0.5969 & \textbf{0.9105} & 0.9006 \\ \hline
\end{tabular}
    }

\end{table*}

%% file: tables/bscan-selection.tex
\begin{table}[!tbh]
    \caption[]{\textbf{Performance of MM-CNN \cite{miccai19-amd} given distinct B-scan selection strategies at the test stage}. Manual selection is needed in order to maximize the performance of MM-CNN.}
    \label{tab:bscan-selection}
    \centering
    \resizebox{\columnwidth}{!}
    {
\begin{tabular}{|l|c|c|c|c|c|}
\hline
\textbf{Which B-scan?} & \textbf{AP} & \textbf{AUC} & \textbf{Sen.} & \textbf{Spe.} & \textbf{F1} \\ \hline
First-frame & 0.7551 & 0.9081 & 0.7199 & 0.9498 & 0.8078 \\ \hline
Middle-frame & 0.7524 & 0.9168 & 0.6490 & 0.9495 & 0.7448 \\ \hline
Last-frame & 0.7502 & \textbf{0.9188} & 0.6941 & \textbf{0.9532} & 0.7904 \\ \hline
Manual & \textbf{0.7743} & 0.9072 & \textbf{0.7244} & 0.9522 & \textbf{0.8128} \\ \hline
\end{tabular}
    }
\end{table}

%% file: tables/over-sample.tex
\begin{table}[]
\caption{Performance of MM-MIL$\times 1$ with and without over sampling.}
\label{tab:os}
\resizebox{\linewidth}{!}
{
\begin{tabular}{|l|c|c|c|c|c|}
\hline
\textbf{Over sampling?} & \textbf{AP} & \textbf{AUC} & \textbf{Sen.} & \textbf{Spe.} & \textbf{F1} \\ \hline
Yes & \textbf{0.8097} & 0.9261 & \textbf{0.7311} & \textbf{0.9635} & \textbf{0.8214} \\ \hline
No & 0.7872 & \textbf{0.9286} & 0.6729 & 0.9468 & 0.7640 \\ \hline
\end{tabular}
}
\end{table}

%% file: tables/mhsa.tex
\begin{table}[H]
    \caption{Performance of our multi-modal network with distinct fusion modules, \ie MM-MIL versus Multi-Head self-Attention (MHA). Our \emph{MM-MIL} achieves better performance with much fewer parameters (see Table \ref{tab:mmmil_vs_mha}).}
    \label{tab:mhsa}
    \resizebox{\linewidth}{!}{
        \begin{tabular}{|l|c|c|c|c|c|}
        \hline
        \textbf{Fusion module} &  \textbf{AP} & \textbf{AUC} & \textbf{Sen.} & \textbf{Spe.} & \textbf{F1} \\ \hline
        MHA (d=1, h=1) &  0.8116 & 0.9442 & 0.7055 & \textbf{0.9668} & 0.8061 \\ \hline
        MHA (d=4, h=4) &  0.8043 & 0.9310 & 0.6982 & 0.9589 & 0.8025 \\ \hline
        \emph{MM-MIL}$\times 4$ &  \textbf{0.8539} & \textbf{0.9545} & \textbf{0.7431} & 0.9643 & \textbf{0.8291} \\ \hline
        \end{tabular}
    }
\end{table}